\documentclass[journal]{IEEEtran}
\usepackage{hyperref} 
\usepackage{subfloat}
\usepackage{subcaption}
\usepackage[pdftex]{graphicx}
\usepackage{graphics}

\usepackage{booktabs}
\usepackage{multirow}
\usepackage{longtable}
\usepackage{mwe}
\usepackage{times}
\usepackage{epsfig}
\usepackage{booktabs}
\usepackage[dvipsnames]{xcolor}
\usepackage{xspace}
\usepackage{pifont}
\let\oldCopyright\copyright
\let\oldTextregistered\textregistered
\usepackage{float}
\usepackage{rotating}
\newlength{\tempdima}
\newcommand{\rowname}[1]
{\rotatebox{90}{\makebox[\tempdima][c]{\normalsize{#1}}}}
\renewcommand{\copyright}{\textsuperscript{\oldCopyright}\xspace}
\renewcommand{\textregistered}{\textsuperscript{\oldTextregistered}\xspace}

\graphicspath{{../images/}}
\DeclareGraphicsExtensions{.pdf,.jpeg,.png}
\usepackage{amsmath}
\usepackage{amssymb}
\usepackage{amsfonts}
\usepackage{array}
\usepackage{url}
\usepackage{color, soul, colortbl}

\usepackage{caption}
\newcommand{\beginsupplement}{%
        \setcounter{table}{0}
        \renewcommand{\thetable}{S\arabic{table}}%
        \setcounter{figure}{0}
        \renewcommand{\thefigure}{S\arabic{figure}}%
     }

\begin{document}

\title{Automated dermatoscopic pattern discovery by clustering neural network output for human-computer interaction}

\author{
    \IEEEauthorblockN{Lidia Talavera-Martinez\IEEEauthorrefmark{1}, Philipp Tschandl\IEEEauthorrefmark{2}}\\
    \IEEEauthorblockA{\IEEEauthorrefmark{1}SCOPIA Research Group, University of the Balearic Islands}\\
    \IEEEauthorblockA{\IEEEauthorrefmark{2}Department of Dermatology, Medical University of Vienna, Austria}
}

\maketitle
\begin{abstract}
\ul{Background:} As available medical image datasets increase in size, it becomes infeasible for clinicians to review content manually for knowledge extraction. The objective of this study was to create an automated clustering resulting in human-interpretable pattern discovery. 

\ul{Methods:} Images from the public HAM10000 dataset, including 7 common pigmented skin lesion diagnoses, were tiled into 29420 tiles and clustered via k-means using neural network-extracted image features. The final number of clusters per diagnosis was chosen by either the elbow method or a compactness metric balancing intra-lesion variance and cluster numbers. The amount of resulting non-informative clusters, defined as those containing less than six image tiles, was compared between the two methods.

\ul{Results:} Applying k-means, the optimal elbow cutoff resulted in a mean of 24.7 (95\%-CI: 16.4-33) clusters for every included diagnosis, including 14.9\% (95\% CI: 0.8-29.0) non-informative clusters. The optimal cutoff, as estimated by the compactness metric, resulted in significantly fewer clusters (13.4; 95\%-CI 11.8-15.1; p=0.03) and less non-informative ones (7.5\%; 95\% CI: 0-19.5; p=0.017). The majority of clusters (93.6\%) from the compactness metric could be manually mapped to previously described dermatoscopic diagnostic patterns.

\ul{Conclusions:} Automatically constraining unsupervised clustering can produce an automated extraction of diagnostically relevant and human-interpretable clusters of visual patterns from a large image dataset.

\end{abstract}

Pre-peer review version \footnote{This is the pre-peer reviewed version of the following article: \emph{Talavera-Martinez L, Tschandl P. Automated dermatoscopic pattern discovery by clustering neural network output for human-computer interaction. J Eur Acad Dermatol Venereol. 2023}, which has been published in final form at \url{https://doi.org/10.1111/jdv.19234}. This article may be used for non-commercial purposes in accordance with Wiley Terms and Conditions for Use of Self-Archived Versions.}

\section{Introduction}

\IEEEPARstart{I}{n} dermatology, but also other visual medical fields, the recognition and description of specific samples of diseases is important for a precise diagnosis and for the formulation of differential diagnoses. Apart from clinical dermatology, there have been a plethora of pattern-descriptions of symptoms of disease in dermatoscopy in recent decades \cite{kittler2016standardization}, especially for the diagnosis of skin tumors and inflammatory diseases \cite{errichetti2020standardization}, which are used for teaching and diagnosis in daily practice. These descriptions were mostly based on mono- or multicentric case collections that were reviewed manually by a few authors and evaluated on possible repetitions of patterns \cite{menzies2000surface,braun2002dermoscopy,cameron2010dermatoscopy}. As clinical image data collections are increasing in size \cite{han2018classification,combalia2019bcn20000}, entirely manual review for discovering diagnostic patterns is not a realistic scenario anymore. In addition to an insurmountable workload, interrater disagreement may be a hindering factor in identifying and describing objective, valid and teachable pattern groups \cite{kittler2016standardization}.

Increasingly, neural networks - especially convolutional neural networks (CNN) - are described as an aid for diagnostic classification of medical images. In the field of dermatology, CNNs were described to have at least an equal accuracy as dermatologists in experimental settings for classifying clinical and dermatoscopic images, and shown to improve physicians’ diagnostic accuracy when applied in diverse interactive settings \cite{tschandl2020human,han2020augmented}. Such algorithms can not only classify images but also label anatomic areas \cite{amruthalingam2022improved}, rate psoriasis \cite{okamoto2022artificial}, or retrieve similar images to a case, by implicitly analyzing patterns and pattern combinations after training to categorize images into distinct classes \cite{tschandl2019diagnostic}. Therefore, we hypothesize that convolutional neural networks could be helpful in the extraction of diagnostically relevant patterns in medical image collections. Recent reports have shown unsupervised techniques when having limited label data \cite{krammer2022deep}. 

\begin{figure}[ht!]
\centering
\includegraphics[width=1.0\columnwidth]{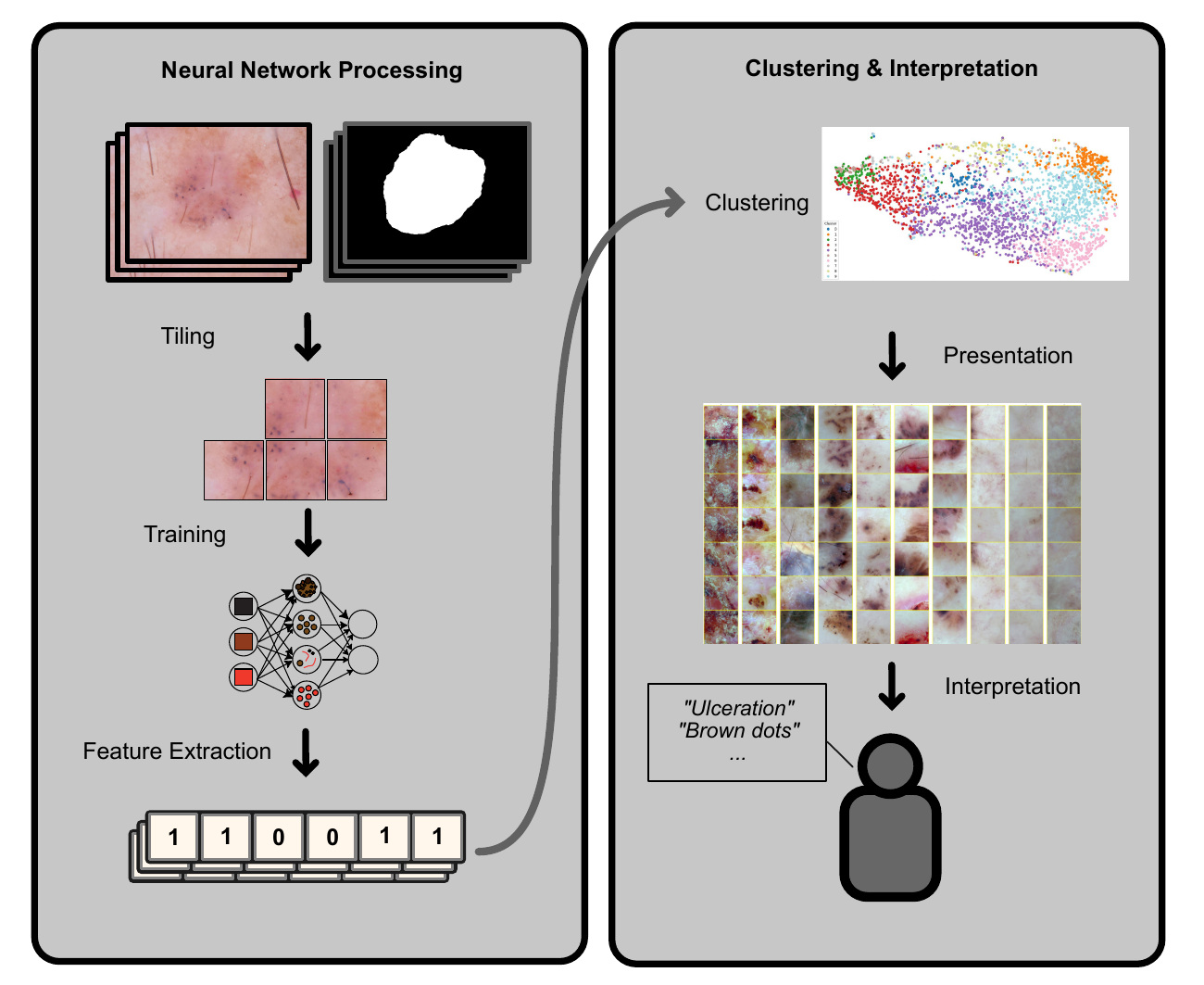}
\caption{Processing overview - Dermatoscopic images (upper left) are used as source data, and a neural network is trained for classification on tiled lesion-area tiles. Features are extracted from lesion tiles with this trained network, on which an unsupervised k-means clustering is applied to find pattern groups. Up to the closest 7 lesion-tiles within a cluster (examples shown for clusters within the BCC class) are stored as representatives of a pattern and presented to a human reader for qualitative interpretation.}
\label{fig:1}
\end{figure}

The goal of this study was to create an automated workflow to extract diagnostic relevant pattern candidates for review by doctors and researchers, with dermatoscopic images of skin tumors as an example (Fig. \ref{fig:1}). Eventually, from a big dataset with thousands of images, this should enable human-computer interaction and return an interpretable number of visually distinct patterns, by obtaining as few redundant or uninformative patterns as possible.

The approach we propose is a pipeline based on machine learning that consists of extracting deep features from CNNs, and applying an unsupervised clustering algorithm to these features. The clustering shall be constrained by a custom compactness metric that, in contrast to the well known elbow-metric, should better balance retrieval of all relevant patterns while at the same time keeping redundant information low.

\section{Materials and Methods}

\subsection{Data and processing}
\begin{figure}[ht!]
\centering
\resizebox{5cm}{!} {
\includegraphics[width=\linewidth]{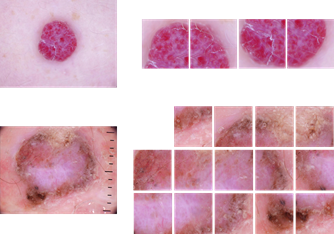}
}
\caption{Example of the tiles extracted (right) from the images of the HAM10000 dataset (left) following the sliding window strategy and that compose our dataset.} \label{suppfig:tileextract}
\end{figure}

\begin{figure}[ht!]
\centering
\resizebox{5cm}{!} {
\includegraphics[width=\linewidth]{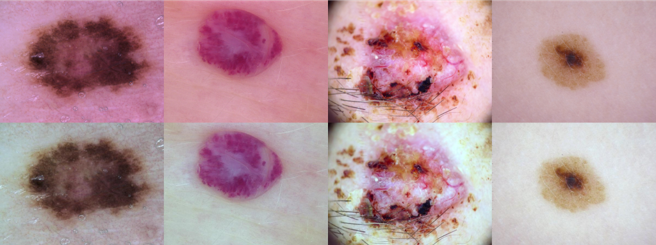}
}
\caption{Top row shows original images from the HAM10000 dataset, and the bottom row results after color normalization.} \label{suppfig:colorNorm}
\end{figure}

This non-interventional retrospective study was conducted on public image data only, specifically the HAM10000 dataset \cite{tschandl2018ham10000}. This dataset is composed of 10015 dermatoscopic images of pigmented lesions with annotations on both the diagnosis and segmentation of the lesion area \cite{tschandl2020human}. To focus on patterns rather than full images, analyses were performed on a tile-level. We extracted square subregions (tiles) of an image by a sliding window with a size of 128x128 pixels with 25\% overlap, discarding tiles with $<60\%$ lesion area. In sum, 29420 tiles were extracted, Suppl. Fig. \ref{suppfig:tileextract} shows two example cases with resulting extracted tiles, and Suppl. Table \ref{supptab:image_to_tiles} the number of tiles per diagnosis. To ensure approximately equal representation of diagnoses, included nevi were limited to a random subsample of 1100 cases, and resulting tiles limited to a maximum random subsample of 850 tiles. To reduce the influence of changes in illumination color, we applied color constancy correction \cite{barata2014improving} to all tiles (Suppl. Fig. \ref{suppfig:colorNorm}).

\subsection{Neural network and feature extraction}
A VGG16 \cite{simonyan2014very} architecture, pretrained on ImageNet data, was fine-tuned to classify tiles into one of seven diagnoses included in the HAM10000 dataset. Training was performed with all 29420 tiles, using 70\% for training and the remaining 30\% for validation during a single training run, ensuring no overlap of tiles of the same image between sets. This training run was only performed as a means to parameterize the model, thus as knowledge discovery rather than classification accuracy was the goal, the complete training dataset was also used for cluster analyses downstream. Data augmentation steps were flips in both horizontal and vertical directions, random 90º rotations, and zooms. Training was performed with a batch size of 32, using the Adam \cite{kingma2014adam} optimizer, a weighted categorical cross entropy loss, with an initial learning rate of 1-e5, and an early stopping policy based on validation loss. For extracting features from image tiles by the fine-tuned model, the numerical state of the layer before the classification layer was obtained, resulting in a 1280-length vector. Neural network experiments were conducted using tensorflow \cite{abadi2016tensorflow} and python 3.8. Experiments were repeated with EfficientNet-B0 \cite{tan2019efficientnet} and a convolutional autoencoder, with results for those two models shown in the supplementary data. For the autoencoder, we trained the model from scratch with a mean-squared error loss, and extracted the features from the flattened embedding space.

\subsection{Clustering}
Resulting extracted features are normalized and used as input to an unsupervised clustering algorithm, specifically k-means \cite{hartigan1979algorithm} with cosine distance as a distance metric. This calculation was performed using scikit-learn v1.1.2 \cite{pedregosa2011scikit} and scipy v1.9.0 \cite{virtanen2020scipy}. 
To automatically obtain the optimal number of clusters without further intervention from a user, either the elbow method (optimal value as calculated by yellowbrick v1.5 \cite{bengfort2019yellowbrick}) or a custom compactness metric (W) was applied. The latter method is based on the assumption that each lesion, thus also the tiles that comprise it, on average only show one or two dermatoscopic patterns. Thus, the proposed metric measures both the similarity of the clusters to which tiles of the same image have been assigned, and the number of different clusters the tiles were assigned to in respect to the total number of clusters. The metric was implemented as follows:

\begin{equation*}
     I=\left\{ img_1,..., img_q \right\}
\end{equation*}
\begin{equation*}
     T_q=\left\{ t_1,..., t_j \right\}
\end{equation*}
\begin{equation*}
    C_q=\left\{ c_1,..., c_i \right\}
\end{equation*}
\begin{multline*}
    W=\frac{1}{M}\times
    \sum_{q=1}^{M}\left ( \frac{K}{min(n_{clst}, L)} \times \sum_{L}^{j=1} cosDst \left ( \frac{1}{K}\sum_{i=1}^{K}(c_i),t_j \right )\right )
\end{multline*}

where $M$ is the number of images $I$ in the experiment, $T_q$ are the $L$ tiles of an image $img_q$, $K$ is the number of unique clusters to which $T_q$ belong, $n_{clst}$ are the total number of clusters used in the experiment. Reiterating, the first half of W for an image ensures tiles are spread to as few clusters as possible, and the second half of W ensures the distance of tiles to the common center of clusters, that tiles are assigned to, is low.

\subsection{Classification ability}
To assess classification ability of the two clustering cutoff methods, clusters were created not only for each diagnosis separately, but also for the whole dataset spanning all diagnoses. The frequency of diagnoses contained in a resulting cluster was noted as a multi-class probability for a classification task. Test images from the ISIC 2018 challenge Task 3 \cite{tschandl2019comparison,codella2019skin} were tiled and preprocessed as above (resulting in 10254 tiles from 1304 lesions with sufficient lesion area depicted), and probabilities of the closest cluster of each tile averaged. The top-1 class of the resulting probabilities was taken as a prediction, and accuracy as well as mean recall \cite{tschandl2019comparison} calculated.

\subsection{Manual pattern descriptions}
Top-7 tiles of clusters, created for every diagnosis in the dataset separately with VGG16 feature vectors, k-means and the described compactness metric, were inspected by a dermatologist with substantial experience in dermatoscopy (PT). Patterns were scored for redundancy, i.e. showing the same pattern as another cluster of the diagnosis, informativeness, i.e. whether any reproducible pattern can be identified, number of patterns, and previous description, i.e. whether the pattern was already identified and described in the literature. A pattern was defined as a change in color and/or structure covering the majority of the image tile.

\subsection{Statistics}
Differences of paired values were compared using a one-sample t-test after checking for normality assumptions. Statistical analyses were performed using R Statistics v4.1.0 \cite{r2014r}, and plots created with ggplot2 \cite{wickham2016data}. A two sided p-value $<.05$ was regarded as statistically significant, with a Bonferroni-Holm type correction applied.

\section{Results}

\subsection{Pattern interpretability}
\begin{figure}[ht!]
\centering
\begin{tabular}{c}
    \begin{subfigure}[b]{0.5\columnwidth}
         \centering
         \includegraphics[width=\columnwidth]{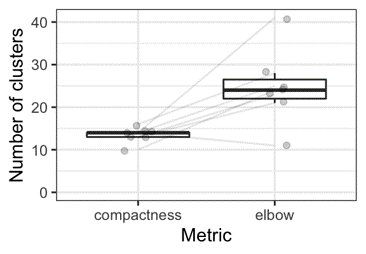}
         \caption{}
         \label{fig:2A}
     \end{subfigure}
     
    \begin{subfigure}[b]{0.5\columnwidth}
         \centering

         \includegraphics[width=\columnwidth]{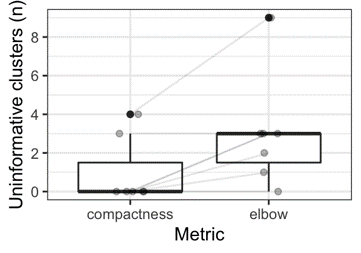}
         \caption{}
         \label{fig:2B}
     \end{subfigure}
\end{tabular}
\caption{Number of overall (a) and uninformative (b) clusters per diagnosis when constraining cluster numbers via either the compactness or elbow method.}
\label{fig:2}
\end{figure}

Applied on clusters of every diagnosis separately, the elbow method created a mean of 24.7 (95\%-CI: 16.4-33) clusters for a diagnosis, whereas the compactness metric resulted in significantly less clusters (13.4; 95\%-CI 11.8-15.1; p=0.03; Fig. \ref{fig:2A}). The proportion of uninformative clusters was higher when using the elbow method (14.9\%; 95\% CI: 0.8-29.0) than using the compactness metric (7.5\%, 95\% CI: 0-19.5; p=0.017; Fig. \ref{fig:2B}).

Qualitative interpretation of the clusters resulting from the compactness metric, for 93.6\% (88 of 94) of diagnosis-specific clusters, at least one recognizable consistent pattern could be identified by a dermatologist. Identified patterns could be mapped to 53 unique known diagnostic descriptions from previous literature of at least 29 publications (Suppl. Table \ref{supptab:qualitative_Annotation}). Only 51 clusters could be described with one pattern alone, whereas 30 clusters encompassed two, and 7 clusters three recognizable patterns in combination. The proportion of redundant clusters within a diagnosis ranged from 0\% (basal cell carcinoma and melanoma) to 27.3\% (dermatofibroma and vascular lesions).

\subsection{Retained classification performance}
\begin{figure}[ht!]
\centering
\begin{tabular}{c}
    \begin{subfigure}[b]{0.5\columnwidth}
         \centering
         \includegraphics[width=\textwidth]{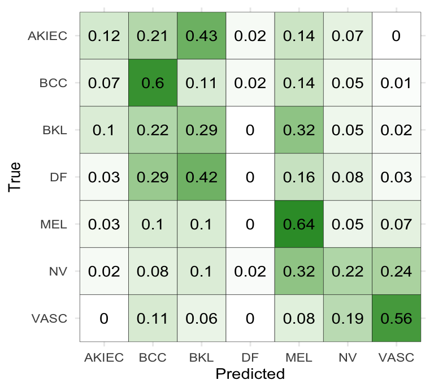}
         \caption{}
         \label{fig:3A}
     \end{subfigure}
     
    \begin{subfigure}[b]{0.5\columnwidth}
         \centering
         \includegraphics[width=\textwidth]{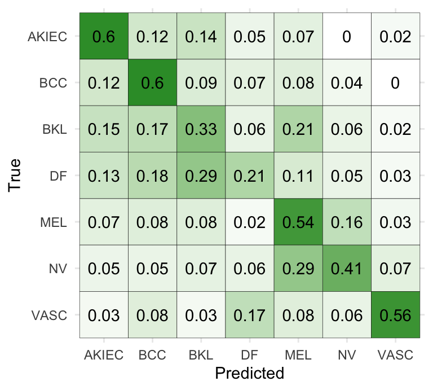}
         \caption{}
         \label{fig:3B}
     \end{subfigure}
\end{tabular}
\caption{Confusion matrices showcasing performance of predictions via averaged nearest neighbor cluster-probabilities constrained by (a) compactness metric (7 clusters), or (b) the elbow method (42 clusters). Values within cells show proportions within one ground-truth class (=row).}
\label{fig:3}
\end{figure}

Application of clustering on the whole dataset with all diagnoses included, using the elbow method resulted in a higher number of clusters than the compactness metric (42 vs. 7), as well as a higher mean recall (46.3 vs. 34.6) and accuracy (43.4\%; 95\%-CI 40.7-46.2 vs. 32.2\%; 95\%-CI 29.7-34.8) for predictions on the ISIC2018 test set. Clusters of the compactness metric were rarely able to predict actinic keratoses, and almost never dermatofibroma (Fig. \ref{fig:3}).

\section{Discussion}
With ever growing image datasets, human interpretation of available data becomes increasingly difficult, and herein we present an automated analysis pipeline which via automated image processing is supposed to aid human-computer interaction for diagnostic marker discovery. Providing only information on the diagnosis and image area, we were able to showcase that the presented workflow is able to reproduce a major fraction of diagnostic patterns in dermatoscopy described in the literature.

In contrast to other publications \cite{han2018classification,tschandl2019comparison,haenssle2020man} trying to optimize for the best diagnostic accuracy of a neural network model, herein we propose a metric to constrain k-means clustering to optimize for human interpretability in a truly interactive human-computer interaction workflow. The proposed compactness metric reduces the information to a digestible amount, shown by the significant reduction in overall clusters (Fig \ref{fig:2A}), alongside a reduction of non-informative information shown by the significant reduction of noninformative clusters (Fig \ref{fig:2B}). These improvements, though come at a cost, namely a reduced diagnostic accuracy when applied in an automated classification setting. This underlines that the training proposed herein could be useful for human-computer interaction and interpretability, but not for safely predicting diagnoses as a standalone application. As biases from automated predictions of image data through neural networks is a significant problem \cite{groh2021evaluating}, datasets should be inspected for potential biases. Although not explicitly shown in this pilot experiment, the proposed workflow may enable medical personnel and researchers to identify highly prevalent biases in a qualitative manner. It is certainly not a complete solution, as based on the failure to classify rare classes (Fig \ref{fig:3A}) we hypothesize that biases on rare classes will equally not be detectable.

Through qualitative analysis of resulting clusters we found that for most it is not possible to find a single pattern to describe them, but the majority needed at least a combination of two patterns (Supp. Table \ref{supptab:qualitative_Annotation}). This finding may help in designing future annotation and pattern analysis studies, and we hypothesize that studies trying to annotate and analyze for a single structure may not be representing real patterns. Interestingly, this may be a missing link between descriptive and “metaphoric” language \cite{kittler2016standardization}, as the former is more suitable for distinct and concise descriptions, but metaphoric language inherently tries to capture structure combinations. A further interesting insight was that the frequency of redundant clusters was not equally distributed, but higher in dermatofibroma and vascular lesions. This could be sourced by the fact that these diagnoses in general show less variability in their patterns, but also that the used dataset through small sample size for these diagnoses is not covering the real visual variability. Finally, it is also interesting to note that by qualitatively comparing different network architectures (Supp. Fig. \ref{supptab:akiec_fig} - \ref{supptab:vasc_fig}), one can identify different utility of them for the purpose of pattern discovery. While an Autoencoder is mainly detecting color blobs, edges, corners and curves, it is focussing less on detailed structures. The top-7 tiles from clusters created by using EfficientNetB0, as a representative of a modern architecture with higher diagnostic accuracy than VGG16 \cite{tan2019efficientnet}, were less homogeneous and thus harder to interpret. Thus, despite being not ideal for classification, we hypothesize that VGG16, through its inner architecture, is a well fit for extracting features of  interpretable mid-level patterns useful for human-computer interaction. Future studies should show feasibility of implementation of this workflow not only for present dermatoscopic datasets \cite{zaballos2008dermoscopy}, but also other imaging modalities such as dermatopathology and clinical images.

\subsection{Limitations}
This pilot study was supposed to showcase the general feasibility of the proposed process. Applicability to nonpigmented tumors, other localisations, inflammatory cases and darker skin types cannot be estimated as those were not included in the source datasets. The process at its core is analyzing substructures of dermatoscopic images, thus overall architecture is not integrated but could theoretically be overcome by changing the tile size and minimal lesion area. The latter is a relevant consideration when applying the workflow, as with the initially chosen tile size and lesion area constraints, some test cases with a very small lesion depicted did not produce any tile.

\section*{Acknowledgements}
Lidia Talavera-Mart\'inez was a beneficiary of the scholarship BES-2017-081264 granted by the Ministry of Economy, Industry, and Competitiveness of Spain under a program co-financed by the European Social Fund.  She is also part of the R\&D\&i  Project PID2020-113870GB-I00, funded by MCIN/AEI/10.13039/50110
0011033/.

\section*{Data availability} 
Used image data are openly available at \url{https://doi.org/10.7910/DVN/DBW86T} (Harvard Dataverse). Resulting Clusters and qualitative evaluations are available in the supplementary material of this article.


{\small
\bibliographystyle{IEEEtran}
\bibliography{egbib}
}


\newpage
\onecolumn

\vspace*{\fill}
\LARGE
\centerline{Supplementary Data}
\vspace*{\fill}

\newpage
\normalsize 
\beginsupplement

\begin{table}[!ht]
\centering
\begin{tabular}{|c|c|c|}
\hline
\textbf{Diagnosis}           & \textbf{Nº images} & \textbf{Nº Tiles} \\ \hline
{AKIEC} & 327                & 2993              \\ \hline
{BCC}   & 514                & 2989              \\ \hline
{BKL}   & 1099               & 7992              \\ \hline
{DF}    & 115                & 825               \\ \hline
{MEL}   & 1113               & 8262              \\ \hline
{NV}    & 6705               & 5927              \\ \hline
{VASC}  & 142                & 432               \\ \hline
\end{tabular}
\caption{Correspondence between the number of images of the HAM10000 dataset according to diagnosis with the number of tiles extracted and used during the experimentation.}
\label{supptab:image_to_tiles}
\end{table}

\begin{table}[!ht]
\centering
\begin{tabular}{ll|c|c|}
\cline{3-4}
\multicolumn{2}{l|}{\textbf{}} & \textbf{Elbow} & \textbf{Compactness} \\ \hline
\multicolumn{1}{|l|}{{Diagnosis}} & Feature extraction model & optimal nº of clusters & optimal nº of clusters \\ \hline
\multicolumn{1}{|l|}{} & EfficientNet B0 & 23 & 18 \\ \cline{2-4} 
\multicolumn{1}{|l|}{} & VGG16 & 23 & 13 \\ \cline{2-4} 
\multicolumn{1}{|l|}{\multirow{-3}{*}{{AKIEC}}} & AE & 38 & 12 \\ \hline
\multicolumn{1}{|l|}{} & EfficientNet B0 & 17 & 14 \\ \cline{2-4} 
\multicolumn{1}{|l|}{} & VGG16 & 24 & 10 \\ \cline{2-4} 
\multicolumn{1}{|l|}{\multirow{-3}{*}{{BCC}}} & AE & 34 & 15 \\ \hline
\multicolumn{1}{|l|}{} & EfficientNet B0 & 21 & 15 \\ \cline{2-4} 
\multicolumn{1}{|l|}{} & VGG16 & 28 & 16 \\ \cline{2-4} 
\multicolumn{1}{|l|}{\multirow{-3}{*}{BKL}} & AE & 33 & 18 \\ \hline
\multicolumn{1}{|l|}{} & EfficientNet B0 & 17 & 10 \\ \cline{2-4} 
\multicolumn{1}{|l|}{} & VGG16 & 11 & 14 \\ \cline{2-4} 
\multicolumn{1}{|l|}{\multirow{-3}{*}{DF}} & AE & 24 & 14 \\ \hline
\multicolumn{1}{|l|}{} & EfficientNet B0 & 40 & 16 \\ \cline{2-4} 
\multicolumn{1}{|l|}{} & VGG16 & 25 & 14 \\ \cline{2-4} 
\multicolumn{1}{|l|}{\multirow{-3}{*}{MEL}} & AE & 45 & 13 \\ \hline
\multicolumn{1}{|l|}{} & EfficientNet B0 & 35 & 12 \\ \cline{2-4} 
\multicolumn{1}{|l|}{} & VGG16 & 41 & 14 \\ \cline{2-4} 
\multicolumn{1}{|l|}{\multirow{-3}{*}{NV}} & AE & 49 & 13 \\ \hline
\multicolumn{1}{|l|}{} & EfficientNet B0 & 15 & 18 \\ \cline{2-4} 
\multicolumn{1}{|l|}{} & VGG16 & 21 & 13 \\ \cline{2-4} 
\multicolumn{1}{|l|}{\multirow{-3}{*}{VASC}} & AE & 24 & 20 \\ \hline
\multicolumn{1}{|l|}{} & EfficientNet B0 & 28 & 10 \\ \cline{2-4} 
\multicolumn{1}{|l|}{} & VGG16 & 42 & 7 \\ \cline{2-4} 
\multicolumn{1}{|l|}{\multirow{-3}{*}{All}} & AE & 34 & 6 \\ \hline
\end{tabular}
\caption{Resulting cluster numbers by diagnosis, model and clustering cutoff method. W denotes the compactness value.}
\label{supptab:clusternum_by_diagnosis_and_method}
\end{table}

\tiny
\begin{longtable}{l c c l l l c l c}
\hline
\textbf{Diagnosis} & \textbf{\begin{tabular}[c]{@{}c@{}}Cluster num.\\ (arbitrary)\end{tabular}} & \textbf{Inconclusive} & \multicolumn{1}{c|}{\textbf{Pattern 1}} & \multicolumn{1}{c|}{\textbf{Pattern 2}} & \multicolumn{1}{c|}{\textbf{Pattern 3}} & \textbf{Redundant to} & \multicolumn{1}{c|}{\textbf{\begin{tabular}[c]{@{}c@{}}Term from\\ previous literature\end{tabular}}} & \textbf{Literature} \\ \hline \hline
\endfirsthead
\hline
\hline
\textbf{Diagnosis} & \textbf{\begin{tabular}[c]{@{}c@{}}Cluster num.\\ (arbitrary)\end{tabular}} & \textbf{Inconclusive} & \multicolumn{1}{c|}{\textbf{Pattern 1}} & \multicolumn{1}{c|}{\textbf{Pattern 2}} & \multicolumn{1}{c|}{\textbf{Pattern 3}} & \textbf{Redundant to} & \multicolumn{1}{c|}{\textbf{\begin{tabular}[c]{@{}c@{}}Term from\\ previous literature\end{tabular}}} & \textbf{Literature} \\ \hline \hline
\endhead
\multicolumn{2}{c}{Continues on next page.}
\endfoot
\endlastfoot

{} & 1 & 0 & \begin{tabular}[c]{@{}l@{}}dots, grey,\\ linear\end{tabular} & \begin{tabular}[c]{@{}l@{}}vessels, red,\\ linear\end{tabular} &  &  & Dots-as-lines & {\cite{cameron2010dermatoscopy}} \\ \cline{2-9} 
{} & 6 & 0 & \begin{tabular}[c]{@{}l@{}}circles,\\ white\end{tabular} & \begin{tabular}[c]{@{}l@{}}lines,\\ grey/brown,\\ angulated\end{tabular} &  &  & Angulated lines & {\cite{tschandl2015dermatoscopy}} \\ \cline{2-9} 
{} & 12 & 0 & \begin{tabular}[c]{@{}l@{}}structureless,\\ brown\end{tabular} &  &  &  & Structureless brown & {\cite{inskip2020dermatoscopic}} \\ \cline{2-9} 
{} & 9 & 0 & circles, white & \begin{tabular}[c]{@{}l@{}}structureless,\\ brown\end{tabular} &  &  & White circles & {\cite{inskip2020dermatoscopic}} \\ \cline{2-9} 
{} & 2 & 0 & \begin{tabular}[c]{@{}l@{}}follicles, skin-\\ colored,\\ prominent\end{tabular} & \begin{tabular}[c]{@{}l@{}}structureless,\\ brown\end{tabular} &  &  & Prominent follicles & {\cite{peris2007dermoscopic}} \\ \cline{2-9} 
{} & 11 & 0 & \begin{tabular}[c]{@{}l@{}}structureless,\\ white\end{tabular} &  &  &  & Structureless white & {\cite{cameron2010dermatoscopy}} \\ \cline{2-9} 
{} & 5 & 0 & circles, white & \begin{tabular}[c]{@{}l@{}}structureless,\\ brown\end{tabular} & hair, black, thick & 9 & White circles & {\cite{tschandl2015dermatoscopy}} \\ \cline{2-9} 
{} & 3 & 0 & \begin{tabular}[c]{@{}l@{}}follicles, skin-\\ colored,\\ prominent\end{tabular} & \begin{tabular}[c]{@{}l@{}}structureless,\\ brown, angulated\end{tabular} & dots, grey &  & Prominent follicles & {\cite{tschandl2015dermatoscopy}} \\ \cline{2-9} 
\multirow{-13}{*}{{\textbf{AKIEC}}} & 8 & 0 & \begin{tabular}[c]{@{}l@{}}structureless,\\ brown\end{tabular} & \begin{tabular}[c]{@{}l@{}}telangiectasia,\\ red, blurry\end{tabular} &  &  & Linear-wavy vessels & {\cite{zalaudek2006dermoscopy}} \\ \cline{2-9} 
{} & 7 & 0 & \begin{tabular}[c]{@{}l@{}}structureless,\\ brown\end{tabular} & dots, brown &  &  & Brown dots and structureless & {\cite{cameron2010dermatoscopy}} \\ \cline{2-9} 
{} & 4 & 0 & \begin{tabular}[c]{@{}l@{}}structureless,\\ brown, linear\end{tabular} &  &  &  & - &  \\ \cline{2-9} 
{} & 10 & 0 & \begin{tabular}[c]{@{}l@{}}structureless,\\ brown, clustered\end{tabular} &  &  &  & Structureless brown, clusters & {\cite{cameron2010dermatoscopy}} \\ \cline{2-9} 
 & 0 & 0 & \begin{tabular}[c]{@{}l@{}}dots, brown,\\ linear\end{tabular} &  &  & 1 & Dots-as-lines & {\cite{cameron2010dermatoscopy}} \\ \hline
{} & 3 & 0 & \begin{tabular}[c]{@{}l@{}}structures,\\ white\end{tabular} & \begin{tabular}[c]{@{}l@{}}telangiectasia,\\ red, blurry\end{tabular} &  &  & Shiny white structures & {\cite{scalvenzi2008dermoscopic}} \\ \cline{2-9} 
{} & 2 & 0 & \begin{tabular}[c]{@{}l@{}}structureless,\\ skin-colored\end{tabular} & \begin{tabular}[c]{@{}l@{}}telangiectasia,\\ red, blurry\end{tabular} &  &  & Shiny white structures & {\cite{scalvenzi2008dermoscopic}} \\ \cline{2-9} 
{} & 4 & 0 & \begin{tabular}[c]{@{}l@{}}structureless,\\ pink\end{tabular} & \begin{tabular}[c]{@{}l@{}}reticular lines,\\ brown, delicate\end{tabular} &  &  & Network-like structures & {\cite{gulia2010pigmented}} \\ \cline{2-9} 
{} & 7 & 0 & \begin{tabular}[c]{@{}l@{}}clod, brown,\\ radial/peripheral\\ (?leaf like?)\end{tabular} &  &  &  & Leaf-like & {\cite{puspok1997statistical}} \\ \cline{2-9} 
{} & 6 & 1 & lesion border &  &  &  &  &  \\ \cline{2-9} 
{} & 5 & 0 & \begin{tabular}[c]{@{}l@{}}clod, brown/blue,\\ in another clod\end{tabular} & \begin{tabular}[c]{@{}l@{}}structureless,\\ white/pink,\\ as background\end{tabular} &  &  & Clod within a clod & {\cite{kittler2016standardization}} \\ \cline{2-9} 
{} & 8 & 0 & \begin{tabular}[c]{@{}l@{}}clod, brown/blue,\\ in another clod\end{tabular} & \begin{tabular}[c]{@{}l@{}}structureless,\\ brown, as\\ background\end{tabular} &  &  & Clod within a clod & {\cite{kittler2016standardization}} \\ \cline{2-9} 
\multirow{-10}{*}{{\textbf{BCC}}}  & 1 & 1 & \begin{tabular}[c]{@{}l@{}}structureless,\\ purple\end{tabular} &  &  &  & Blue-gray globules & {\cite{peris2002interobserver}} \\ \cline{2-9} 
{} & 9 & 0 & clod, red/orange & \begin{tabular}[c]{@{}l@{}}structureless,\\ pink\end{tabular} & structures, white &  & Ulceration & {\cite{peris2002interobserver}} \\ \cline{2-9} 
& 0 & 0 & \begin{tabular}[c]{@{}l@{}}structureless,\\ white/orange/red\end{tabular} &  &  &  & Ulceration & {\cite{peris2002interobserver}} \\ \hline
{} & 6 & 0 & \begin{tabular}[c]{@{}l@{}}structureless,\\ brown/grey,\\ speckled\end{tabular} &  &  &  & Granular pattern & {\cite{bugatti2007dermoscopy}} \\ \cline{2-9} 
{} & 15 & 0 & \begin{tabular}[c]{@{}l@{}}structureless,\\ brown/grey\end{tabular} &  &  &  & Blotch & {\cite{braun2002dermoscopy}} \\ \cline{2-9} 
{} & 5 & 0 & \begin{tabular}[c]{@{}l@{}}structureless,\\ brown/red\end{tabular} & \begin{tabular}[c]{@{}l@{}}clods,\\ orange/skin-colored\end{tabular} &  &  & Comedo-like openings & {\cite{braun2002dermoscopy}} \\ \cline{2-9} 
{} & 8 & 0 & \begin{tabular}[c]{@{}l@{}}structureless,\\ brown/grey\end{tabular} &  &  &  & Blotch & {\cite{braun2002dermoscopy}} \\ \cline{2-9} 
{} & 10 & 0 & \begin{tabular}[c]{@{}l@{}}structureless,\\ brown\end{tabular} & \begin{tabular}[c]{@{}l@{}}lesion border,\\ sharp demarcation\end{tabular} &  &  & Sharp demarcation & {\cite{braun2002dermoscopy}} \\ \cline{2-9} 
{} & 11 & 0 & \begin{tabular}[c]{@{}l@{}}structureless,\\ brown, speckled\end{tabular} & \begin{tabular}[c]{@{}l@{}}reticular lines,\\ brown\end{tabular} &  &  & ?False? reticular lines & {\cite{de2002false}} \\ \cline{2-9} 
{} & 1 & 0 & dots, grey &  &  &  & Coarse granules & {\cite{zaballos2007dermoscopic}} \\ \cline{2-9} 
{} & 2 & 0 & dots, brown & \begin{tabular}[c]{@{}l@{}}structureless,\\ brown\end{tabular} &  &  & Coarse granules & {\cite{zaballos2007dermoscopic}} \\ \cline{2-9} 
 & 13 & 0 & \begin{tabular}[c]{@{}l@{}}reticular lines,\\ brown, thick\end{tabular} &  &  &  & Thickened network & {\cite{braun2002dermoscopy}} \\ \cline{2-9} 
\multirow{-16}{*}{{\textbf{BKL}}} & 3 & 0 & \begin{tabular}[c]{@{}l@{}}clods, purple,\\ with central vessel\end{tabular} &  &  &  & Exophytic papillary structure & {\cite{braun2002dermoscopy}} \\ \cline{2-9} 
{} & 14 & 0 & \begin{tabular}[c]{@{}l@{}}structureless,\\ light brown\end{tabular} &  &  &  & Blotch & {\cite{braun2002dermoscopy}} \\ \cline{2-9} 
& 12 & 0 & \begin{tabular}[c]{@{}l@{}}curved lines,\\ brown, thick\end{tabular} &  &  &  & fat fingers & {\cite{kopf2006fat}} \\ \cline{2-9} 
{} & 4 & 1 & circles, brown &  &  &  & - &  \\ \cline{2-9} 
{} & 9 & 0 & \begin{tabular}[c]{@{}l@{}}structureless,\\ brown\end{tabular} & \begin{tabular}[c]{@{}l@{}}reticular lines,\\ brown, uneven\end{tabular} &  & 11 & ?False? reticular lines & {\cite{de2002false}} \\ \cline{2-9} 
{} & 7 & 0 & \begin{tabular}[c]{@{}l@{}}structureless,\\ brown\end{tabular} & \begin{tabular}[c]{@{}l@{}}dots, brown,\\ single\end{tabular} &  &  & Granular pattern & {\cite{bugatti2007dermoscopy}} \\ \cline{2-9} 
 & 0 & 0 & \begin{tabular}[c]{@{}l@{}}structureless,\\ blue/grey\end{tabular} &  &  &  & Blotch & {\cite{braun2002dermoscopy}} \\ \hline
{} & 3 & 0 & \begin{tabular}[c]{@{}l@{}}structureless,\\ brown\end{tabular} & \begin{tabular}[c]{@{}l@{}}structureless,\\ white, central\end{tabular} &  &  & Central white scarlike tile & {\cite{ferrari2000central}} \\ \cline{2-9} 
{} & 4 & 0 & \begin{tabular}[c]{@{}l@{}}structureless,\\ red, prominent\\ skin markings\end{tabular} &  &  &  & Reddish coloration & {\cite{ferrari2000central}} \\ \cline{2-9} 
{} & 11 & 0 & \begin{tabular}[c]{@{}l@{}}structureless,\\ brown, central\\ white/red\end{tabular} &  &  & 3 & Central white scarlike tile & {\cite{ferrari2000central}} \\ \cline{2-9} 
{} & 6 & 0 & \begin{tabular}[c]{@{}l@{}}structureless, \\ red/skin-colored\end{tabular} & \begin{tabular}[c]{@{}l@{}}scaling, white,\\ delicate\end{tabular} &  &  & Reddish coloration & {\cite{ferrari2000central}} \\ \cline{2-9} 
{} & 1 & 0 & \begin{tabular}[c]{@{}l@{}}reticular lines,\\ brown, delicate\\ and broken up\end{tabular} &  &  &  & Pigment network & {\cite{ferrari2000central}} \\ \cline{2-9} 
{} & 8 & 1 & - &  &  &  & - &  \\ \cline{2-9} 
{} & 9 & 0 & structureless, red & \begin{tabular}[c]{@{}l@{}}scaling,\\ white/orange,\\ dense/peripheral\\ circular\end{tabular} &  &  & Scale crusts & {\cite{ferrari2000central}} \\ \cline{2-9} 
{} & 13 & 1 & - &  &  &  & - &  \\ \cline{2-9} 
{} & 5 & 0 & \begin{tabular}[c]{@{}l@{}}structureless,\\ brown\end{tabular} & \begin{tabular}[c]{@{}l@{}}structureless,\\ white/pink, in corner\end{tabular} &  & 3 & Central white scarlike tile & {\cite{ferrari2000central}} \\ \cline{2-9} 
\multirow{-14}{*}{{\textbf{DF}}}  & 10 & 0 & \begin{tabular}[c]{@{}l@{}}structureless,\\ brown/skin-colored\end{tabular} &  &  &  & - &  \\ \cline{2-9} 
{} & 7 & 0 & clods, brown &  &  &  & Brown globules & {\cite{ferrari2000central}} \\ \cline{2-9} 
{} & 12 & 0 & clods, brown &  &  & 7 & Brown globules & {\cite{ferrari2000central}} \\ \cline{2-9} 
{} & 2 & 0 & \begin{tabular}[c]{@{}l@{}}structureless,\\ white/pink,\\ parallel linear\\ arrangement\end{tabular} &  &  &  & - &  \\ \cline{2-9} 
& 0 & 1 & - &  &  &  & - &  \\ \hline
 & 7 & 0 & \begin{tabular}[c]{@{}l@{}}structureless,\\ brown/grey\end{tabular} &  &  &  & Grey structures & {\cite{tschandl2015dermatoscopy}} \\ \cline{2-9} 
 & 1 & 0 & \begin{tabular}[c]{@{}l@{}}structureless,\\ brown\end{tabular} &  &  &  & Light brown structureless areas & {\cite{annessi2007sensitivity}} \\ \cline{2-9} 
 & 4 & 0 & \begin{tabular}[c]{@{}l@{}}structureless,\\ skin-colored\end{tabular} & \begin{tabular}[c]{@{}l@{}}telegeangiectasia,\\ pink, blurry\end{tabular} &  &  & Depigmentation & {\cite{pizzichetta2001dermoscopic}} \\ \cline{2-9} 
 & 9 & 0 & \begin{tabular}[c]{@{}l@{}}reticular lines,\\ brown\end{tabular} & \begin{tabular}[c]{@{}l@{}}structureless,\\ brown\end{tabular} & lines, white &  & White scar-like areas & {\cite{pizzichetta2001dermoscopic}} \\ \cline{2-9} 
 & 11 & 0 & \begin{tabular}[c]{@{}l@{}}structureless,\\ black/grey\end{tabular} &  &  &  & Blue-black pigmentation & {\cite{rodriguez2021dermoscopic}} \\ \cline{2-9} 
 & 13 & 0 & \begin{tabular}[c]{@{}l@{}}colors,\\ brown/black/white\end{tabular} &  &  &  & Blue-white veil & {\cite{argenziano1998epiluminescence}} \\ \cline{2-9} 
 & 6 & 0 & \begin{tabular}[c]{@{}l@{}}reticular lines,\\ brown, uneven\end{tabular} & dots, grey & \begin{tabular}[c]{@{}l@{}}lesion border, \\ scalloped\end{tabular} &  & Regression & {\cite{argenziano1998epiluminescence}} \\ \cline{2-9} 
 & 12 & 0 & \begin{tabular}[c]{@{}l@{}}structureless, brown,\\ uneven distribution\end{tabular} &  &  &  & Nonuniform pigment distribution & {\cite{annessi2007sensitivity}} \\ \cline{2-9} 
 & 10 & 0 & \begin{tabular}[c]{@{}l@{}}structureless,\\ blue\end{tabular} & \begin{tabular}[c]{@{}l@{}}dots, brown,\\ single\end{tabular} &  &  & Streaks & {\cite{argenziano1998epiluminescence}} \\ \cline{2-9} 
 & 5 & 0 & \begin{tabular}[c]{@{}l@{}}reticular lines,\\ brown, focal\\ hyperpigmentation\end{tabular} &  &  &  & Mistletoe sign & {\cite{kaminska2013mistletoe}} \\ \cline{2-9} 
\multirow{-14}{*}{\textbf{MEL}} & 2 & 0 & \begin{tabular}[c]{@{}l@{}}lines, brown,\\ angulated\end{tabular} &  &  &  & Angulated lines & {\cite{jaimes2015clinical}} \\ \cline{2-9} 
 & 8 & 0 & \begin{tabular}[c]{@{}l@{}}lines, brown,\\ radial/peripheral\end{tabular} &  &  &  & Irregular extensions & {\cite{soyer1995diagnostic}} \\ \cline{2-9} 
 & 3 & 1 & \begin{tabular}[c]{@{}l@{}}reticular lines,\\ brown, thick\end{tabular} &  &  &  & - &  \\ \cline{2-9} 
 & 0 & 0 & \begin{tabular}[c]{@{}l@{}}reticular lines,\\ brown, uneven in\\ structure and color\end{tabular} &  &  &  & Irregular pigment network & {\cite{soyer1995diagnostic}} \\ \hline
{} & 10 & 0 & \begin{tabular}[c]{@{}l@{}}structureless,\\ light brown\end{tabular} &  &  &  & Homogeneous pattern & {\cite{hofmann2001dermoscopic}} \\ \cline{2-9} 
{} & 3 & 0 & \begin{tabular}[c]{@{}l@{}}reticular lines,\\ brown, smooth\\ border transition\end{tabular} & \begin{tabular}[c]{@{}l@{}}structureless,\\ brown\end{tabular} & \begin{tabular}[c]{@{}l@{}}follicle openings,\\ skin colored\end{tabular} &  & Typical pigment network & {\cite{argenziano2003dermoscopy}} \\ \cline{2-9} 
{} & 1 & 0 & \begin{tabular}[c]{@{}l@{}}reticular lines,\\ brown, smooth\\ border transition\end{tabular} & \begin{tabular}[c]{@{}l@{}}structureless,\\ brown\end{tabular} &  & 3 & Typical pigment network & {\cite{argenziano2003dermoscopy}} \\ \cline{2-9} 
{} & 13 & 0 & \begin{tabular}[c]{@{}l@{}}reticular lines,\\ brown, smooth\\ border transition\end{tabular} & \begin{tabular}[c]{@{}l@{}}structureless,\\ brown\end{tabular} &  & 3 & Typical pigment network & {\cite{argenziano2003dermoscopy}} \\ \cline{2-9} 
{} & 11 & 0 & \begin{tabular}[c]{@{}l@{}}reticular lines,\\ brown\end{tabular} & \begin{tabular}[c]{@{}l@{}}clods, brown,\\ single\end{tabular} &  &  & Brown globules & {\cite{argenziano2003dermoscopy}} \\ \cline{2-9} 
{} & 2 & 0 & \begin{tabular}[c]{@{}l@{}}reticular lines,\\ grey, delicate\end{tabular} & \begin{tabular}[c]{@{}l@{}}structureless,\\ grey\end{tabular} &  &  & - &  \\ \cline{2-9} 
{} & 9 & 0 & \begin{tabular}[c]{@{}l@{}}clods, brown/skin-\\ colored, with central\\ vessel\end{tabular} & \begin{tabular}[c]{@{}l@{}}scaling, white,\\ delicate\end{tabular} &  &  & Globular pattern & {\cite{argenziano2003dermoscopy}} \\ \cline{2-9} 
{} & 7 & 0 & \begin{tabular}[c]{@{}l@{}}structureless, brown,\\ hypopigmented areas\end{tabular} &  &  &  & White scar-like areas & {\cite{pizzichetta2001dermoscopic}} \\ \cline{2-9} 
{} & 12 & 0 & \begin{tabular}[c]{@{}l@{}}reticular lines,\\ brown/black\end{tabular} &  &  &  & Superficial black network & {\cite{argenziano2003dermoscopy}} \\ \cline{2-9} 
{} & 4 & 0 & \begin{tabular}[c]{@{}l@{}}lesion border, brown,\\ eccentric hyperpigmen-\\ tation\end{tabular} &  &  &  & Mistletoe sign & {\cite{kaminska2013mistletoe}} \\ \cline{2-9} 
\multirow{-14}{*}{{\textbf{NV}}}  & 5 & 1 & \begin{tabular}[c]{@{}l@{}}structureless, brown,\\ hypopigmented areas\end{tabular} & clods, brown &  &  & - &  \\ \cline{2-9} 
{} & 6 & 0 & \begin{tabular}[c]{@{}l@{}}reticular lines, brown,\\ uneven in structure\\ and color\end{tabular} & \begin{tabular}[c]{@{}l@{}}lesion border, \\ scalloped\end{tabular} &  &  & Irregular pigment network & {\cite{soyer1995diagnostic}} \\ \cline{2-9} 
{} & 8 & 0 & \begin{tabular}[c]{@{}l@{}}reticular lines, brown,\\ uneven in structure\end{tabular} &  &  & 6 & Irregular pigment network & {\cite{soyer1995diagnostic}} \\ \cline{2-9} 
 & 0 & 1 & - &  &  &  & - &  \\ \hline
{} & 11 & 0 & clods, blue &  &  &  & Red-blue lacunae & {\cite{gao2021dermoscopic}} \\ \cline{2-9} 
{} & 1 & 0 & clods, red, clustered &  &  &  & Red globules & {\cite{gao2021dermoscopic}} \\ \cline{2-9} 
{} & 2 & 0 & \begin{tabular}[c]{@{}l@{}}structureless, red, skin-\\ colored septae\end{tabular} &  &  &  & White rail & {\cite{zaballos2006dermoscopic}} \\ \cline{2-9} 
{} & 8 & 0 & \begin{tabular}[c]{@{}l@{}}structureless, red, skin-\\ colored septae\end{tabular} &  &  & 2 & White rail & {\cite{zaballos2006dermoscopic}} \\ \cline{2-9} 
{} & 5 & 0 & clods, red, not in focus &  &  &  & Red lacunae & {\cite{gao2021dermoscopic}} \\ \cline{2-9} 
{} & 12 & 0 & clods, red, not in focus &  &  & 5 & Red lacunae & {\cite{gao2021dermoscopic}} \\ \cline{2-9} 
{} & 10 & 0 & \begin{tabular}[c]{@{}l@{}}structureless, red, skin-\\ colored septae\end{tabular} &  &  & 2 & White rail & {\cite{zaballos2006dermoscopic}} \\ \cline{2-9} 
{} & 6 & 0 & \begin{tabular}[c]{@{}l@{}}clod, red, cicular skin-\\ colored demarcation\end{tabular} &  &  &  & White collarette & {\cite{zaballos2006dermoscopic}} \\ \cline{2-9} 
{} & 9 & 1 & - &  &  &  & - &  \\ \cline{2-9} 
 & 4 & 1 & - &  &  &  & - &  \\ \cline{2-9} 
\multirow{-13}{*}{{\textbf{VASC}}} & 3 & 0 & clods, purple & \begin{tabular}[c]{@{}l@{}}scaling, white,\\ dense\end{tabular} & \begin{tabular}[c]{@{}l@{}}clods, black,\\ in scaling\\ (?blood clods?)\end{tabular} &  & Hemorrhagic crusts & {\cite{zaballos2007dermoscopy}} \\ \cline{2-9} 
{} & 7 & 0 & structureless, blue &  &  &  & Red-blue lacunae & {\cite{gao2021dermoscopic}} \\ \cline{2-9} 
 & 0 & 0 & \begin{tabular}[c]{@{}l@{}}structureless, blue, skin-\\ colored septae\end{tabular} &  &  &  & Dark lacunae \& Whitish veil & {\cite{zaballos2007dermoscopy}} \\ \hline
\caption{Qualitative annotation data of clusters via top-7 tiles.}
\label{supptab:qualitative_Annotation}
\end{longtable}



\begin{table}[ht!]
\centering
\resizebox{13cm}{!} {
\begin{tabular}{|ll|l|}
\hline
\multicolumn{2}{|c|}{{\textbf{AKIEC}}} & \multicolumn{1}{c|}{\textbf{Clusters}} \\ \hline \hline
\multicolumn{1}{|l|}{\multirow{8}{*}{\rotatebox{90}{VGG16}}} & \rotatebox{90} {Elbow} & \includegraphics[width=0.7\textwidth]{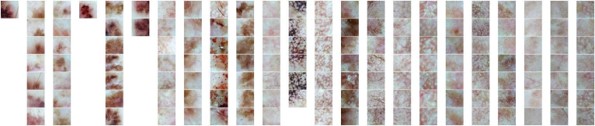}  \\ \cline{2-3} 
 \multicolumn{1}{|l|}{} & \rotatebox{90} {Compactness} & \includegraphics[width=0.7\textwidth]{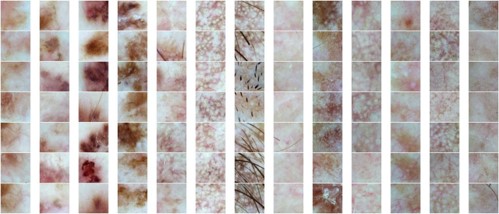}  \\ \hline
\multicolumn{1}{|l|}{\multirow{12}{*}{\rotatebox{90}{Autoencoder}}} & \rotatebox{90} {Elbow} & \includegraphics[width=0.7\textwidth]{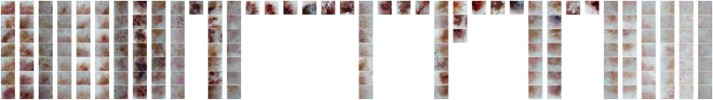}  \\ \cline{2-3} 
\multicolumn{1}{|l|}{} & \rotatebox{90} {Compactness} & \includegraphics[width=0.7\textwidth]{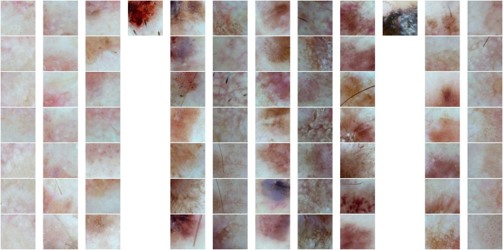}  \\ \hline
\multicolumn{1}{|l|}{\multirow{4}{*}{\rotatebox{90}{EfficientNetB0}}} & \rotatebox{90} {Elbow} & \includegraphics[width=0.7\textwidth]{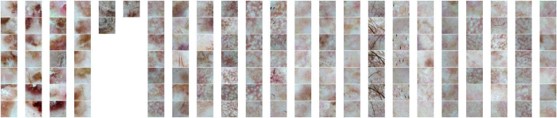}  \\ \cline{2-3} 
\multicolumn{1}{|l|}{} & \rotatebox{90} {Compactness} & \includegraphics[width=0.7\textwidth]{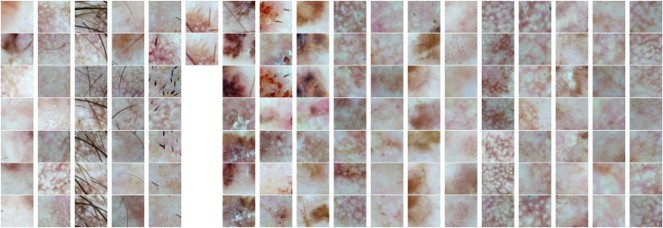}  \\ \hline
\end{tabular}
}
\caption{Clusters resulting from AKIEC diagnosis for each network model (VGG16, Autoencoder, EfficientNetB0) and cluster constraint (Elbow, Compactness). A maximum of 7 tiles closest to the cluster center are shown.}
\label{supptab:akiec_fig}
\end{table}

\begin{table}[h!]
\centering
\resizebox{11.5cm}{!} {
\begin{tabular}{|ll|l|}
\hline
\multicolumn{2}{|c|}{{\textbf{BCC}}} & \multicolumn{1}{c|}{\textbf{Clusters}} \\ \hline \hline
\multicolumn{1}{|l|}{\multirow{12}{*}{\rotatebox{90}{VGG16}}} & \rotatebox{90} {Elbow} & \includegraphics[width=0.7\textwidth]{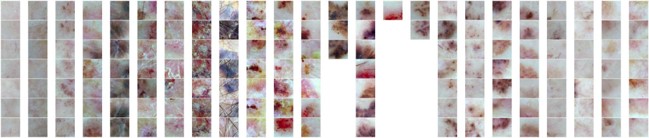}  \\ \cline{2-3} 
 \multicolumn{1}{|l|}{} & \rotatebox{90} {Compactness} & \includegraphics[width=0.7\textwidth]{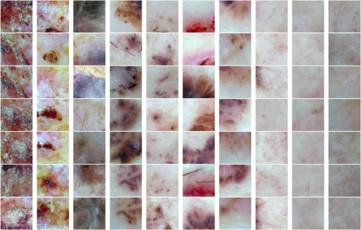}  \\ \hline
\multicolumn{1}{|l|}{\multirow{10}{*}{\rotatebox{90}{Autoencoder}}} & \rotatebox{90} {Elbow} & \includegraphics[width=0.7\textwidth]{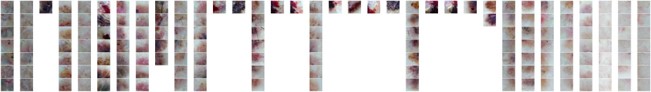}  \\ \cline{2-3} 
\multicolumn{1}{|l|}{} & \rotatebox{90} {Compactness} & \includegraphics[width=0.7\textwidth]{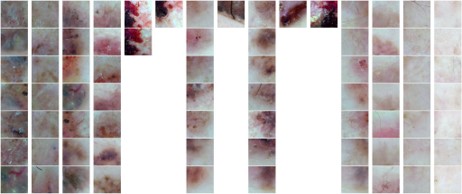}  \\ \hline
\multicolumn{1}{|l|}{\multirow{-2}{*}{\rotatebox{90}{EfficientNetB0}}} & \rotatebox{90} {Elbow} & \includegraphics[width=0.7\textwidth]{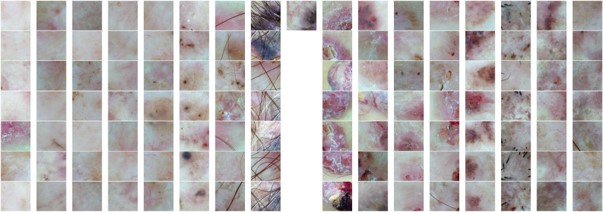}  \\ \cline{2-3} 
\multicolumn{1}{|l|}{} & \rotatebox{90} {Compactness} & \includegraphics[width=0.7\textwidth]{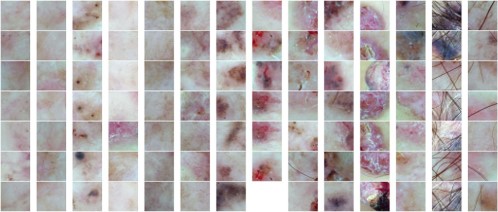}  \\ \hline
\end{tabular}
}
\caption{Clusters resulting from BCC diagnosis for each network model (VGG16, Autoencoder, EfficientNetB0) and cluster constraint (Elbow, Compactness). A maximum of 7 tiles closest to the cluster center are shown.}
\label{suppfig:bcc_fig}
\end{table}

\begin{table}[h!]
\centering
\resizebox{13cm}{!} {
\begin{tabular}{|ll|l|}
\hline
\multicolumn{2}{|c|}{{\textbf{BKL}}} & \multicolumn{1}{c|}{\textbf{Clusters}} \\ \hline \hline
\multicolumn{1}{|l|}{\multirow{6}{*}{\rotatebox{90}{VGG16}}} & \rotatebox{90} {Elbow} & \includegraphics[width=0.7\textwidth]{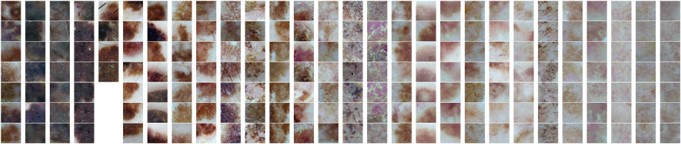}  \\ \cline{2-3} 
 \multicolumn{1}{|l|}{} & \rotatebox{90} {Compactness} & \includegraphics[width=0.7\textwidth]{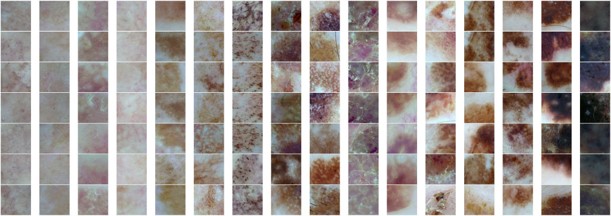}  \\ \hline
\multicolumn{1}{|l|}{\multirow{8}{*}{\rotatebox{90}{Autoencoder}}} & \rotatebox{90} {Elbow} & \includegraphics[width=0.7\textwidth]{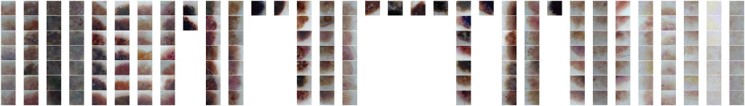}  \\ \cline{2-3} 
\multicolumn{1}{|l|}{} & \rotatebox{90} {Compactness} & \includegraphics[width=0.7\textwidth]{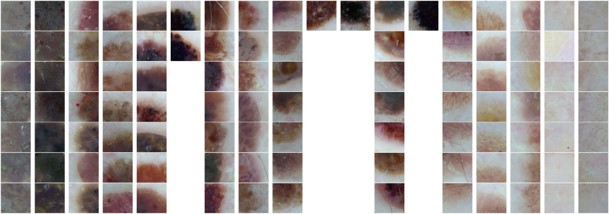}  \\ \hline
\multicolumn{1}{|l|}{\multirow{2}{*}{\rotatebox{90}{EfficientNetB0}}} & \rotatebox{90} {Elbow} & \includegraphics[width=0.7\textwidth]{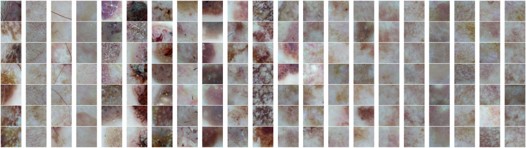}  \\ \cline{2-3} 
\multicolumn{1}{|l|}{} & \rotatebox{90} {Compactness} & \includegraphics[width=0.7\textwidth]{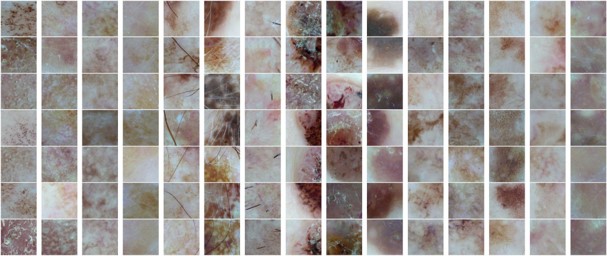}  \\ \hline
\end{tabular}
}
\caption{Clusters resulting from BKL diagnosis for each network model (VGG16, Autoencoder, EfficientNetB0) and cluster constraint (Elbow, Compactness). A maximum of 7 tiles closest to the cluster center are shown.}
\label{supptab:bkl_fig}
\end{table}

\begin{table}[h!]
\centering
\resizebox{10cm}{!} {
\begin{tabular}{|ll|l|}
\hline
\multicolumn{2}{|c|}{{\textbf{DF}}} & \multicolumn{1}{c|}{\textbf{Clusters}} \\ \hline \hline
\multicolumn{1}{|l|}{\multirow{4}{*}{\rotatebox{90}{VGG16}}} & \rotatebox{90} {Elbow} & \includegraphics[width=0.7\textwidth]{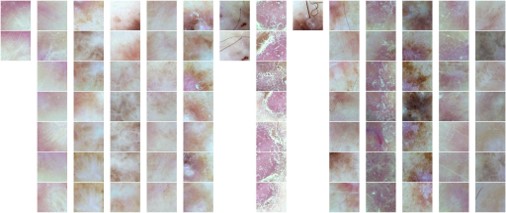}  \\ \cline{2-3} 
 \multicolumn{1}{|l|}{} & \rotatebox{90} {Compactness} & \includegraphics[width=0.7\textwidth]{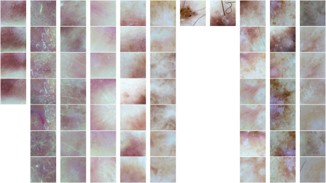}  \\ \hline
\multicolumn{1}{|l|}{\multirow{8}{*}{\rotatebox{90}{Autoencoder}}} & \rotatebox{90} {Elbow} & \includegraphics[width=0.7\textwidth]{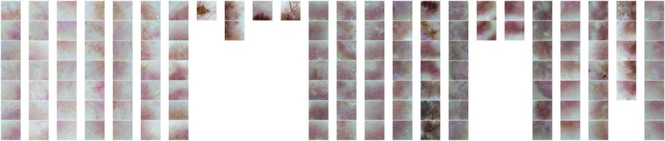}  \\ \cline{2-3} 
\multicolumn{1}{|l|}{} & \rotatebox{90} {Compactness} & \includegraphics[width=0.7\textwidth]{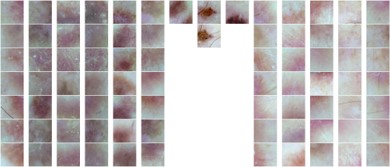}  \\ \hline
\multicolumn{1}{|l|}{\multirow{8}{*}{\rotatebox{90}{EfficientNetB0}}} & \rotatebox{90} {Elbow} & \includegraphics[width=0.7\textwidth]{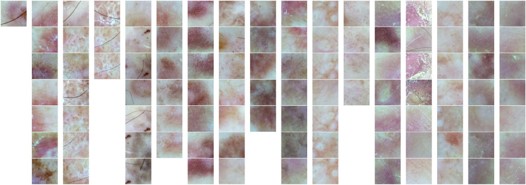}  \\ \cline{2-3} 
\multicolumn{1}{|l|}{} & \rotatebox{90} {Compactness} & \includegraphics[width=0.7\textwidth]{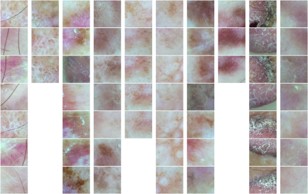}  \\ \hline
\end{tabular}
}
\caption{Clusters resulting from DF diagnosis for each network model (VGG16, Autoencoder, EfficientNetB0) and cluster constraint (Elbow, Compactness). A maximum of 7 tiles closest to the cluster center are shown.}
\label{supptab:DF_fig}
\end{table}

\begin{table}[h!]
\centering
\resizebox{14cm}{!} {
\begin{tabular}{|ll|l|}
\hline
\multicolumn{2}{|c|}{{\textbf{MEL}}} & \multicolumn{1}{c|}{\textbf{Clusters}} \\ \hline \hline
\multicolumn{1}{|l|}{\multirow{8}{*}{\rotatebox{90}{VGG16}}} & \rotatebox{90} {Elbow} & \includegraphics[width=0.7\textwidth]{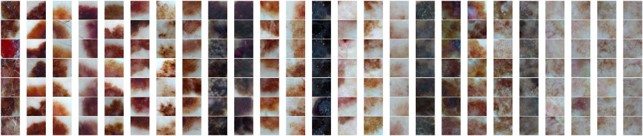}  \\ \cline{2-3} 
 \multicolumn{1}{|l|}{} & \rotatebox{90} {Compactness} & \includegraphics[width=0.7\textwidth]{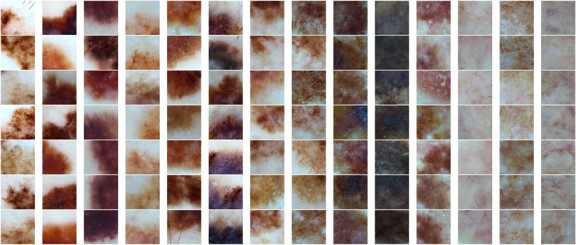}  \\ \hline
\multicolumn{1}{|l|}{\multirow{10}{*}{\rotatebox{90}{Autoencoder}}} & \rotatebox{90} {Elbow} & \includegraphics[width=0.7\textwidth]{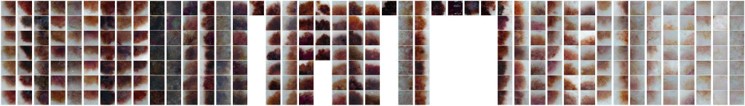}  \\ \cline{2-3} 
\multicolumn{1}{|l|}{} & \rotatebox{90} {Compactness} & \includegraphics[width=0.7\textwidth]{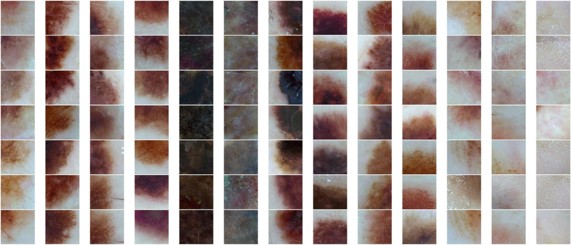}  \\ \hline
\multicolumn{1}{|l|}{\multirow{8}{*}{\rotatebox{90}{EfficientNetB0}}} & \rotatebox{90} {Elbow} & \includegraphics[width=0.7\textwidth]{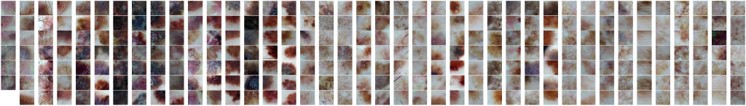}  \\ \cline{2-3} 
\multicolumn{1}{|l|}{} & \rotatebox{90} {Compactness} & \includegraphics[width=0.7\textwidth]{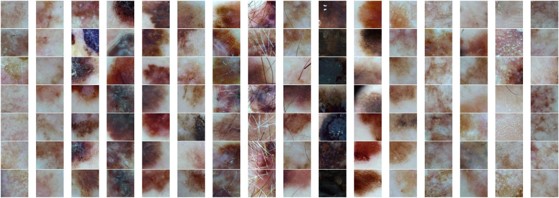}  \\ \hline
\end{tabular}
}
\caption{Clusters resulting from MEL diagnosis for each network model (VGG16, Autoencoder, EfficientNetB0) and cluster constraint (Elbow, Compactness). A maximum of 7 tiles closest to the cluster center are shown.}
\label{supptab:mel_fig}
\end{table}

\begin{table}[h!]
\centering
\resizebox{14cm}{!} {
\begin{tabular}{|ll|l|}
\hline
\multicolumn{2}{|c|}{{\textbf{NV}}} & \multicolumn{1}{c|}{\textbf{Clusters}} \\ \hline \hline
\multicolumn{1}{|l|}{\multirow{10}{*}{\rotatebox{90}{VGG16}}} & \rotatebox{90} {Elbow} & \includegraphics[width=0.7\textwidth]{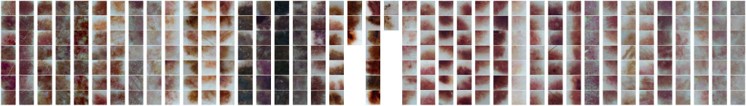}  \\ \cline{2-3} 
 \multicolumn{1}{|l|}{} & \rotatebox{90} {Compactness} & \includegraphics[width=0.7\textwidth]{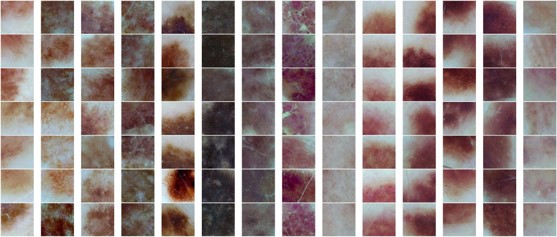}  \\ \hline
\multicolumn{1}{|l|}{\multirow{12}{*}{\rotatebox{90}{Autoencoder}}} & \rotatebox{90} {Elbow} & \includegraphics[width=0.7\textwidth]{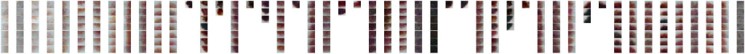}  \\ \cline{2-3} 
\multicolumn{1}{|l|}{} & \rotatebox{90} {Compactness} & \includegraphics[width=0.7\textwidth]{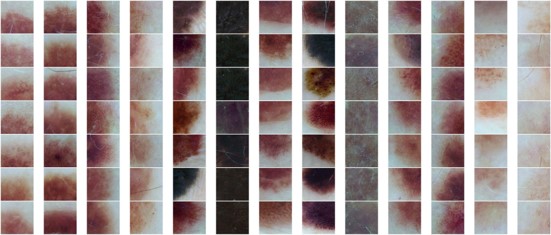}  \\ \hline
\multicolumn{1}{|l|}{\multirow{12}{*}{\rotatebox{90}{EfficientNetB0}}} & \rotatebox{90} {Elbow} & \includegraphics[width=0.7\textwidth]{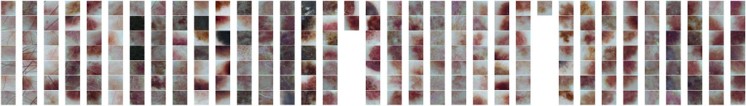}  \\ \cline{2-3} 
\multicolumn{1}{|l|}{} & \rotatebox{90} {Compactness} & \includegraphics[width=0.7\textwidth]{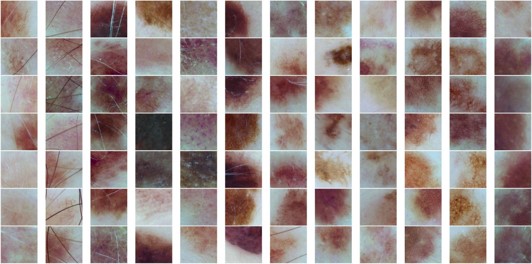}  \\ \hline
\end{tabular}
}
\caption{Clusters resulting from NV diagnosis for each network model (VGG16, Autoencoder, EfficientNetB0) and cluster constraint (Elbow, Compactness). A maximum of 7 tiles closest to the cluster center are shown.}
\label{supptab:nv_fig}
\end{table}

\begin{table}[h!]
\centering
\resizebox{14cm}{!} {
\begin{tabular}{|ll|l|}
\hline
\multicolumn{2}{|c|}{{\textbf{VASC}}} & \multicolumn{1}{c|}{\textbf{Clusters}} \\ \hline \hline
\multicolumn{1}{|l|}{\multirow{6}{*}{\rotatebox{90}{VGG16}}} & \rotatebox{90} {Elbow} & \includegraphics[width=0.7\textwidth]{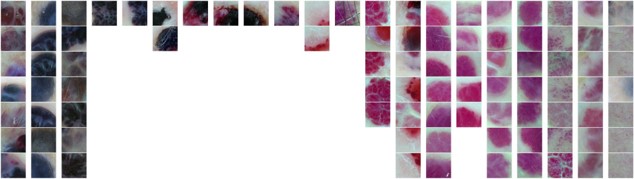}  \\ \cline{2-3} 
 \multicolumn{1}{|l|}{} & \rotatebox{90} {Compactness} & \includegraphics[width=0.7\textwidth]{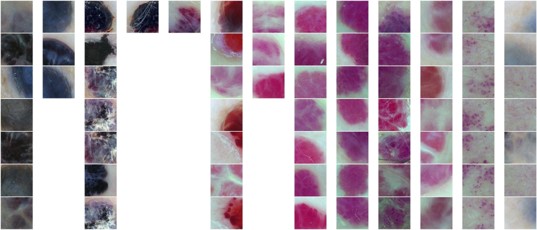}  \\ \hline
\multicolumn{1}{|l|}{\multirow{4}{*}{\rotatebox{90}{Autoencoder}}} & \rotatebox{90} {Elbow} & \includegraphics[width=0.7\textwidth]{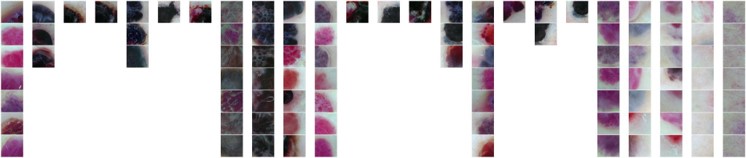}  \\ \cline{2-3} 
\multicolumn{1}{|l|}{} & \rotatebox{90} {Compactness} & \includegraphics[width=0.7\textwidth]{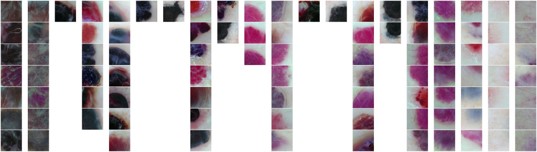}  \\ \hline
\multicolumn{1}{|l|}{\multirow{10}{*}{\rotatebox{90}{EfficientNetB0}}} & \rotatebox{90} {Elbow} & \includegraphics[width=0.7\textwidth]{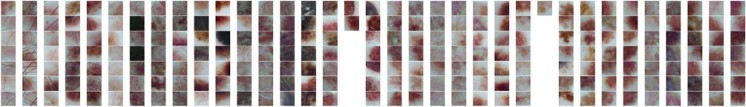}  \\ \cline{2-3} 
\multicolumn{1}{|l|}{} & \rotatebox{90} {Compactness} & \includegraphics[width=0.7\textwidth]{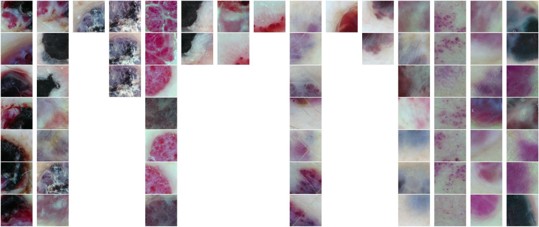}  \\ \hline
\end{tabular}
}

\caption{Clusters resulting from VASC diagnosis for each network model (VGG16, Autoencoder, EfficientNetB0) and cluster constraint (Elbow, Compactness). A maximum of 7 tiles closest to the cluster center are shown.}
\label{supptab:vasc_fig}
\end{table}

\end{document}